\documentclass[10pt,twocolumn,letterpaper]{article}

\usepackage{framed,multirow}

\usepackage{amssymb}
\usepackage{latexsym}
\usepackage{subcaption}
\usepackage{amsmath,graphicx}
\usepackage{amsfonts}
\usepackage{mathtools}
\usepackage{MnSymbol}
\usepackage{lineno}
\usepackage{algorithm}
\usepackage[noend]{algpseudocode}

\usepackage{url}
\usepackage{xcolor}
\definecolor{newcolor}{rgb}{.8,.349,.1}

\usepackage[OMLmathrm,OMLmathbf]{isomath}
\newcommand{\vect}[1]{\mathbf{#1}}

\DeclarePairedDelimiter\ceil{\lceil}{\rceil}

\begin{document}

\title{Ellipsoids from Tangents}

\author{\normalsize 
Efrat Dahan  ~~~~~ Michael Werman\\
\normalsize School of Computer Science \\ 
\normalsize The Hebrew University of Jerusalem, Israel}

\date{}
\maketitle




\begin{abstract}
This paper presents an algorithm for detecting and fitting multiple ellipsoids in 2D, 3D and higher-dimensional data.
The representation is based on tangents to the ellipsoid together with their contact points, rather than on points alone. A tangent contributes $d$ independent equations where a point contributes one, so each hypothesis needs fewer samples, which in turn reduces the number of iterations required to recover every ellipsoid. The problem is posed as sets of linear equations whose solutions --- the ellipsoid parameters --- are recovered robustly by clustering in parameter space. The algorithm works in all dimensions and outperforms the state of the art in 2D. The approach applies in general to the robust solution of linear equations, and so has utility beyond ellipsoid fitting.
\end{abstract}





\section{Introduction}
\label{sec:intro}
Fitting quadrics --- known as conic sections in 2D --- to scattered data in two, three or higher dimensions is a standard pattern recognition problem, and ellipsoids are the most commonly sought quadric. 
Algorithms for fitting ellipses to 2D data can be roughly  split into two approaches. The first works mainly in parameter space, using sets of linear equations usually solved by least squares, which turns ellipse fitting into a constrained optimisation problem \cite{fitzgibbon1999direct, chernov2010circular, rosin1993note,Kanatani20071208,werman1995fitting,masuzaki2013high,Prasad20131449}.
The second approach works mostly in image space, for example edge contour following \cite{prasad2012edge}, which generates many elliptical hypotheses from short local arcs and then groups similar hypotheses \cite{toldo2008robust}. The state of the art for conic fitting \cite{cicconet2014ellipses} exploits both points and their tangents, and proceeds in two steps: first recovering the centres, and only then the remaining parameters.

Algebraic fits are often held to be less accurate than geometric ones, although as \cite{chernov2010circular} observes, that view rests largely on practical experience rather than on analysis. Our results bear on it directly: the algebraic formulation presented here outperforms the geometric state of the art on the same data (Section \ref{sec:exp}).

Methods that work in image space do not generalise easily to 3D or higher dimensions.
The only work we found on  3D problems is the recovery of a single ellipsoid with a known center \cite{bertoni2010multi}.

Both approaches  deal with multiple hypotheses by search in  parameter space with methods such as the  Hough transform \cite{cicconet2014ellipses, chernov2010circular}  or RANSAC \cite{fischler1981random}. Selecting the best hypothesis with a Hough transform is very costly in higher dimensions, while RANSAC is more efficient but does not succeed on very noisy data or on many ellipses, since noise and data from several models can together be counted as evidence for an ellipse that does not exist. 

Most work on 2D ellipse detection uses point data alone. Our method, inspired by \cite{cicconet2014ellipses}, instead uses tangents to recover the ellipsoid's parameters. 
2D tangents can be computed from an image of a digital curve \cite{6873254}, by applying  edge detection  which provides both edges and their directions, e.g., using wavelets as in \cite{bruna2013invariant}, or using a modification of the Canny method, \cite{canny1986computational}.
Planes in higher dimensions can be found with  surface fitting as in \cite{macauslan2010determining} or RANSAC. Tangents, although computed locally, give rich information that can be exploited for  ellipsoid evaluation. 
A likely reason for the scarcity of work on ellipsoid detection in 3D and above is the difficulty of selecting a set of inliers in high dimensions, since the number of data points needed to compute a single ellipsoid grows quadratically with the dimension. This is precisely the cost that tangents reduce.

It is worth saying why we do not simply generalise the geometric voting approach, which is the state of the art in 2D. Two obstacles appear as soon as the dimension rises. Voting for a 2D ellipse already requires more than three points, which is often hard to obtain from real data; and in 3D the natural construction proceeds in two stages --- first voting for the centre of a 2D ellipse, then for the centre of the ellipsoid --- so that the error of the first stage is inherited and amplified by the second. Posing the problem instead as a homogeneous linear system, $H\vect{x} = \vect{0}$, avoids both: the parameters are recovered in one step, from a minimal sample, in any dimension.

\begin{figure*}	
	\centering
	\begin{subfigure}[t]{0.23\textwidth}
		\centering
		\includegraphics[height=3.29cm, width=1.085\linewidth]{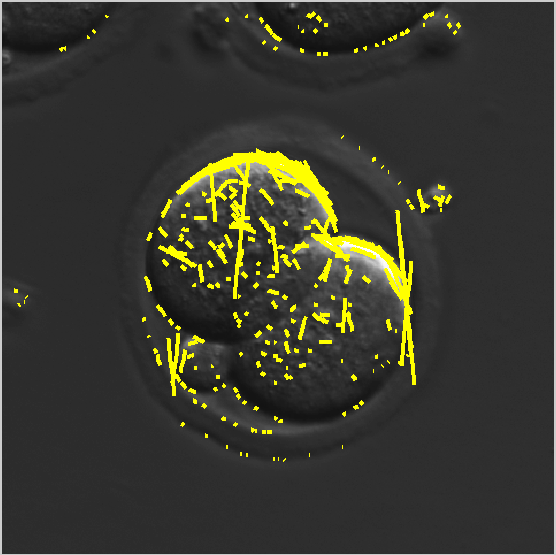}
		\caption{The input, tangents.}\label{fig:1a}
	\end{subfigure}
	\quad
	\begin{subfigure}[t]{0.23\textwidth}
		\centering
		\includegraphics[height=3.29cm, width=1.009\linewidth]{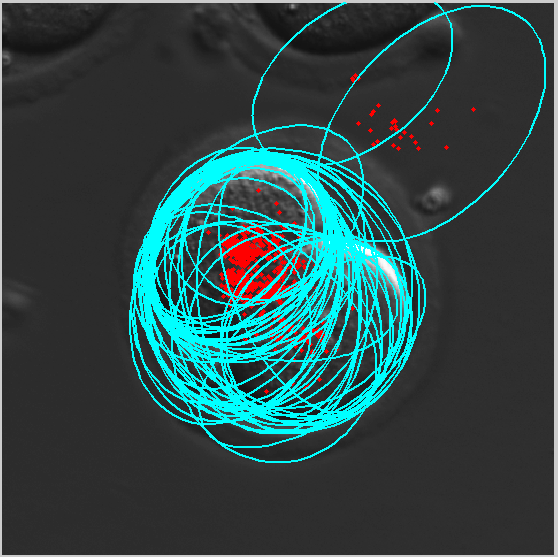}
		\caption{The sampled ellipses.}\label{fig:1b}
	\end{subfigure}
    	\begin{subfigure}[t]{0.23\textwidth}
		\centering
		\includegraphics[height=3.29cm, width=1.085\linewidth]{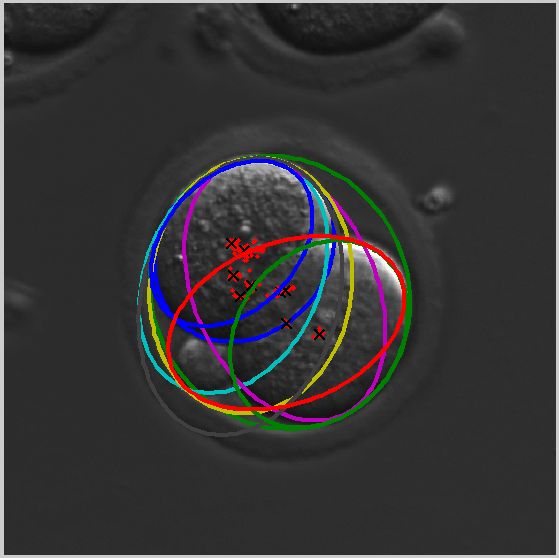}
		\caption{The ellipses that are the centers of the substantial clusters after meanshift clustering.}\label{fig:1c}
	\end{subfigure}
    	\begin{subfigure}[t]{0.23\textwidth}
		\centering
		\includegraphics[height=3.29cm, width=1.085\linewidth]{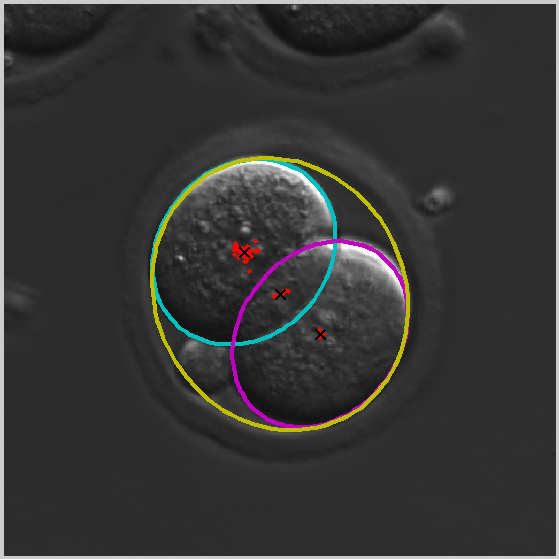}
		\caption{The ellipses left after non maximum suppression.}\label{fig:1d}
	\end{subfigure}
	\caption{The main steps of the algorithm in the 2D case.}\label{fig:2d_ellipse}
\end{figure*}

The algorithm rests on two ideas: tangents and clustering. Computing a quadric from tangents rather than from points (Figure \ref{fig:1a}) allows models to be estimated by linear systems (Figure \ref{fig:1b}). This extends straightforwardly to higher dimensions and, because fewer samples are needed per hypothesis, it raises the probability that a random sample consists entirely of inliers. 
Then meanshift clustering  \cite{fukunaga1975estimation} allows us without knowing the number of clusters  to choose the true models  (Figure \ref{fig:1c}). 
The algorithm works in all dimensions and outperforms state-of-the-art methods.

Nothing in the formulation is specific to ellipsoids. The equations below are written for a general quadric $\vect{A}$, and the quadric type enters only at the filtering step of Section \ref{ssec:normalization}, where solutions of the wrong type are discarded by the signs of the eigenvalues. We nevertheless present everything in terms of ellipsoids: they are the most commonly sought quadric in practice, they keep the exposition concrete, and they are what the experiments measure. The same derivation applies unchanged to the other quadric types, and more generally to any model that can be written as a homogeneous linear system.

The paper is organised as follows. Section \ref{sec:algo}  describes  the  main steps of the algorithm. The algorithm complexity is analyzed in Section \ref{complexity}. Section \ref{sec:exp} presents experiments and comparisons  to other 2D methods, 3D experiments with synthetic data and real models and multidimensional examples with noisy synthetic data.

\section{Tangents}
\label{ssec:elip_as_tan}


Tangents give rich information for ellipsoid evaluation.
Each data point $\vect{p}$ gives a single homogeneous equation,  $\vect{p}^{T}\vect{A}\vect{p} = 0$, for the $\frac{(d+2)(d+1)}{2} - 1$  parameters of an ellipsoid, where $d$ is the dimension and $\vect{A}$ the $(d+1) \times (d+1)$ symmetric matrix representing the ellipse. A tangent, by contrast, gives $d$ independent equations,  $\vect{A}\vect{p} \equiv \vect{t}$, ($\equiv$ being equality up to scale) for    tangent $\vect{t}$ at $\vect{p}$  on  ellipsoid $\vect{A}$, see \cite{hartley2003multiple}.

Basing the algorithm on tangents therefore reduces the number of samples each hypothesis requires, which increases the probability that a sample consists entirely of inliers. Since that probability falls rapidly with the number of samples drawn, the saving compounds: with noise and multiple ellipsoids present, it is the difference between a tractable number of iterations and an intractable one.

For example in 3D: 
The tangent  equations are of form $\vect{A}\vect{p} \equiv \vect{t}$, where $\vect{t}$ is the tangent plane to the ellipsoid $\vect{A}$ at point $\vect{p}$.
$\vect{p} = [x, y, z, 1]$, $\vect{t} = [t_{x}, t_{y}, t_{z}, t_{w}]$.
To convert the equivalence into equalities we normalise as follows:
  ${\vect{A}[1,:]\vect{p}}-\frac{t_{x}}{t_{w}}{\vect{A}[4,:]\vect{p}}=0$, 
${\vect{A}[2,:]\vect{p}}-\frac{t_{y}}{t_{w}}{\vect{A}[4,:]\vect{p}}=0$, and
${\vect{A}[3,:]\vect{p}}-\frac{t_{z}}{t_{w}}{\vect{A}[4,:]\vect{p}}=0$, where  the 10  homogeneous elements of the $4 \times 4$ symmetric matrix
$\vect{A}$ are the unknowns.

For a 3D ellipsoid, three tangents --- each contributing three equations --- suffice to compute the nine normalised parameters (Section \ref{ssec:normalization}). Note that nothing above uses the fact that $\vect{A}$ is an ellipsoid: the same equations hold for any quadric type, which is distinguished only later, when the solutions are filtered.

A numerical solution to this homogeneous system can be found with the SVD decomposition of the linear coefficients matrix, $UDV^{T}$. The column of $V$ corresponding to the smallest singular value gives the non-zero homogeneous solution vector. 


\section{Algorithm}
 \label{sec:algo}
 The algorithm's input is tangents together with their contact points from multiple ellipsoids (and noise) and  outputs  the ellipsoids' parameters.
The algorithm has two main steps --- random sampling and clustering --- and is summarised in Algorithm \ref{alg:main}.
\subsection{Random Sampling - First Step}
\label{ssec:ransac}

An iteration is a random choice of $\kappa$ ($\kappa \geq \frac{\frac{(d+2)(d+1)}{2} - 1} {d}$) tangents. Each $\kappa$ tangents form a linear system whose solution gives the parameters of an  ellipsoid.
The first step ends after $N$ iterations with at most $N$ ellipsoids. $N$, the number of iterations needed, depends on the dimension, the number of ellipsoids and the noise level, and can be computed from the multinomial expression in Section \ref{iterations}.

\subsection{Normalization and Filtering the Solution Vector}
\label{ssec:normalization}
Real-world data is noisy and the model may contain several ellipsoids, so many of the ellipsoids sampled in the previous step are wrong. Taking an additional point, and perhaps an additional tangent, contributes further equations; the cost is that each extra sample increases the number of iterations needed to find all the ellipsoids. When a non-minimal set of equations is used, solutions can be filtered by the smallest singular value, which measures how close the solution is to being a null-space solution: the closer that singular value is to zero, the more accurate the solution, and the more likely it is that the sampled tangents came from a single ellipsoid. 

A second issue is that the coefficient matrix $\vect A$ is a homogeneous representation of an ellipsoid,
so its parameters are defined only up to scale. This is a problem when we wish to compare similar solutions, filter inaccurate ones, or check that parameters are plausible. Normalising the solution vectors makes different solutions comparable for clustering, and allows parameter sets outside known ranges to be discarded. Without it there is nothing meaningful to cluster.
The following lemma gives a normalization factor so that elements of $A$  reflect the real geometry.
\null

\underline{Lemma}:

The geometric scale factor is:
$ g = -\left | \vect A \right | / \left | M \right | $, where M is the principal $d \times d$ minor of $\vect A$.

\underline{Proof}:

$\vect A = g\begin{pmatrix} \vect M & \vect{b}\\ \vect{b}^T & s \end{pmatrix}$
where $\mathbf{b} = \vect{-Mc} \enspace ;
\enspace \vect s = \vect c^{T}\vect M \vect c - 1 $ \enspace ; \enspace $\vect c$ is the center of the ellipsoid. 

$\left | \vect A \right | = g \left |\vect  M  s - {\vect b ^T}\vect M^{-1}\vect{b} \right |$ (see  \cite{silvester2000determinants})
$\Rightarrow \left |\vect  A \right | = g \left |s -  \vect {c^{T}MM^{-1}Mc}\right | \left | \vect  M \right | = g \left | s - \vect{c^{T}Mc}   \right | \left |\vect  M \right | 
\Rightarrow \left| \vect A\right| = -g\left|\vect M\right|$

$
\Rightarrow 
g = -\left |\vect  A \right | / \left |\vect  M \right|  
$
\quad $\blacksquare$

 
\subsection{Clustering - Second Step}
\label{ssec:clustering}

The final step, once the filtered ellipsoids are available as points in parameter space, is to select the most appropriate solutions. Many of the solutions found in the first step tend to cluster around the true ellipsoids, a fact that is also  the basis of RHT \cite{xu1990new}. 

An effective way to cluster is using meanshift \cite{fukunaga1975estimation},
where the cluster centers, ellipsoids' parameters, are the results. An example of  cluster centers is shown in Figure \ref{fig:1c}.
The clusters are then filtered by non-maximum suppression, removing clusters that lie close to a larger one.
An example of the  remaining clusters is shown in Figure \ref{fig:1d}.

\begin{algorithm}[t]
\caption{Fitting multiple ellipsoids from tangents}
\label{alg:main}
\begin{algorithmic}[1]
\Require tangents with their contact points; iteration count $N$; thresholds
\Ensure the ellipsoids' parameters
\State $S \gets \emptyset$ \Comment{candidate solutions in parameter space}
\For{$i = 1$ \textbf{to} $N$}
  \State draw $\kappa$ tangents with their contact points, at random with replacement
  \State solve the homogeneous linear system by SVD
  \State normalise the solution vector (Section \ref{ssec:normalization})
  \State \textbf{if} the smallest singular value is below threshold \textbf{then} add the solution to $S$
\EndFor
\State cluster $S$ by mean shift, keeping clusters of at least $\varsigma$ votes
\State discard clusters close to a larger cluster (non-maximum suppression)
\State \Return the surviving cluster centres
\end{algorithmic}
\end{algorithm}

\section{Complexity}\label{complexity}

\subsection{Number of Iterations}
\label{iterations}
Handling noise and multiple ellipsoids requires repeated sampling of sets of tangents in order to assemble enough correct equations to compute the ellipsoids' parameters.
Here we  bound  the number of iterations required.

We first define the following quantities:

$N$ - the number of iterations.

$\nu$- the total number of tangents.

$\mu$ - the number of ellipsoids.

$\rho_i$ - the number of tangents belonging to the i-th ellipsoid.

$\kappa$ - the number of  tangents needed for one linear system,

$\ceil{ \frac{\frac{(d+2)(d+1)}{2} - 1}d}$.

$\varsigma$ --- the minimal number of votes in a cluster for that cluster to be declared a real ellipsoid.

An iteration is a random choice of $\kappa$ tangents (together with their contact points), the number of tangents sufficient to compute an ellipsoid's parameters, from the data.
To analyze the number of iterations needed we compute the probability to find all the ellipsoids. A run is successful when each of the $\mu$ ellipsoids is sampled at least $\varsigma$ times, the minimal cluster size. 

The probability of drawing, in a single iteration, $\kappa$ points that all belong to the $i$-th ellipsoid is
$prob_i = \frac{\binom{\rho_i}{\kappa}}{\binom{\nu}{\kappa}}$.
The random variable $X_i$ indicates the number of iterations in which all the $\kappa$ sampled tangents are from the i-th ellipsoid, in $N$ trials. The vector $X = (X_1, ..., X_\mu,N-\sum X_i)$ follows a multinomial distribution with parameters $N$ and $P$, where $P = (prob_1, ..., prob_\mu,prob_{\mu+1})$. Notice that $\sum_{i=1}^{\mu}prob_i < 1$ as  tangents can be chosen from multiple ellipsoids.
$\begin{multlined}
\mathbb{P}(X_1 \geqslant \varsigma, X_2 \geqslant \varsigma, \ldots, X_\mu \geqslant \varsigma) = \sum_{\xi_1=\varsigma}^{N}\sum_{\xi_2=\varsigma}^{N} \cdots\\ \sum_{\xi_\mu=\varsigma}^{N} \binom{N}{\xi_1, \xi_2, \ldots , \xi_\mu, N-\sum \xi_i} \\ \times\, prob_{1}^{\xi_1} \cdots prob_{\mu}^{\xi_\mu}\left(1-\sum prob_i\right)^{N-\sum \xi_i} 
\end{multlined}$
For example, reaching a probability above 0.99 of finding two ellipses, each with 100 tangents and at least five samples per cluster, requires 100 iterations; with a further 100 outliers, 400 iterations are needed.
In 3D  the same setup requires   300 iterations for the clean case and 1000 iterations with outliers.

\subsection{Total Number of Operations}
The algorithm has  two main parts, computing the parameters of  ellipsoids and clustering similar solutions to find the real parameters.
The first stage solves $N$ linear systems of size $\frac{(d+2)(d+1)}{2} -1$
   by SVD,  filtering, and normalization. The time spent to compute the solution of a linear equation is cubic in the number of equations.
The second stage applies mean shift to a set of at most $N$ points in $\frac{(d+2)(d+1)}{2}-1$ dimensions with $T$ iterations and is bounded by $O(TN^2d^2)$.
Therefore, overall the algorithm's complexity is $O(TN^2d^2+Nd^6)$.


The running time for 1000 iterations was approximately 0.8~seconds in 2D and 2.3~seconds in 3D, on a laptop with a Core~i3 M370 CPU at 2.40~GHz.

\section{Experiments}
\label{sec:exp}
Much work has been done on 2D ellipse detection, and very little in 3D and above. 

\subsection{2D Experiments}
\begin{table}[h!]
\caption{Results of our method compared with Ellipses From Triangles (EFT) and Prasad's method.   NI  and  NE  stand  for  number  of  images  and  cells, respectively. The “Prasad (Or.)”  rows correspond to the original implementation of the salient elliptic hypotheses method.}
\centering
 {   \begin{tabular}{|| c | c | c | c | c||} 
    \hline
    NI  & NE & Method & Precision & Recall \\ 
    \hline 
     &  & Ours $tolerance < 20$ & 96\% & 95\%   \\  
     &  & Ours $tolerance < 15$ & 93\% & 92\%   \\  
    421& 842 & EFT & 93\% & 92\%  \\ 
    & & Prasad & 77\% & 50\%   \\ 
    & & Prasad (Or.) & 82\% & 13\% \\ 
    \hline
\end{tabular}}

\label{tab:1}
\end{table}

For real 2D images we used two sets of microscopy images of mouse embryos, each containing two cells per image, from the database of \cite{Cicconet2014}. 
These sets contain 421 images each.
We compared our algorithm to other approaches,  \cite{cicconet2014ellipses,prasad2012edge}, on cell embryo images using the quoted results from \cite{cicconet2014ellipses}. 

A detection is considered correct if the recovered centre lies within 15 of the ground truth.
Since the ground-truth values for the 2D embryo images come from a circle-fitting algorithm and contain only the centre, the angles and radii of the computed ellipses were not evaluated. With a larger tolerance of 20 we improve not only recall but also precision. Results are in Table \ref{tab:1}.
\begin{figure}
\begin{minipage}[b]{1.0\linewidth}
  \centering
  \centerline{\includegraphics[width=8.5cm, height=4.5cm]{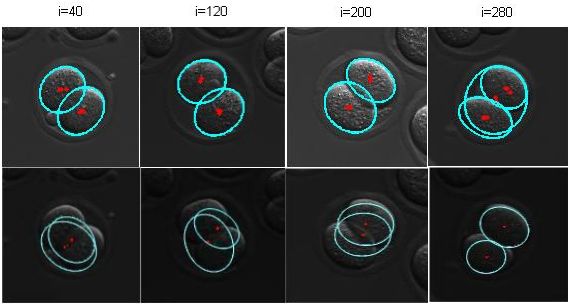}}
\end{minipage}
\caption{Comparing our algorithms (first row) results' parameters against EFT (second row)}\label{fig:comp_efts}
\end{figure}

The experiment above compares centres only, since that is the ground truth available. 
Our ellipses visibly fit the cells better. The frame number is marked above each image in Figure \ref{fig:comp_efts}.
\begin{figure}
\begin{minipage}[b]{1.0\linewidth}
  \centering
  \centerline{\includegraphics[width=8.8cm, height=3.0cm]{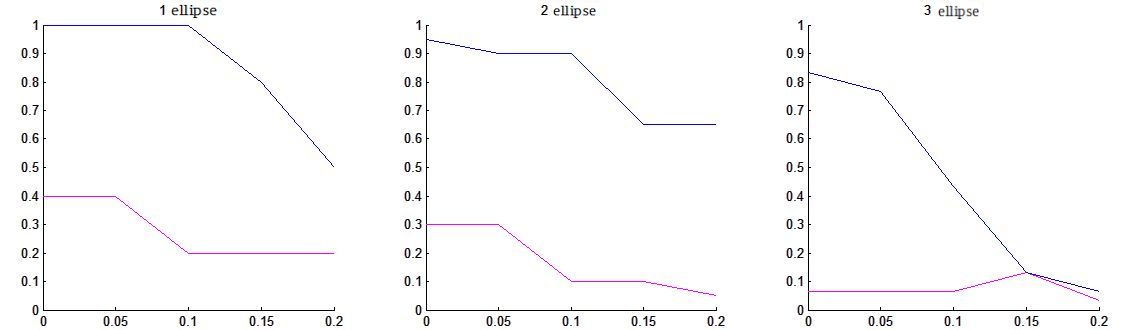}}
\end{minipage}
\caption{The blue line is our algorithm, the red line is the EFT algorithm. The X-axis is the standard deviation of the gaussian noise and the Y-axis is the  success fraction.}\label{fig:syn_2d}
\end{figure}

In order to compare  all the parameters of 2D ellipse fitting, 
we used synthetic images: 150 images of size $350 \times 350$, containing one, two or three ellipses each, with the points and tangents perturbed by Gaussian noise. A trial counts as a success when all parameters fall within 15 of their true values. The outcomes for one, two and three ellipses are shown in Figure \ref{fig:syn_2d}.

A further experiment tests the last step of our algorithm, mean-shift clustering, against RANSAC. We took 30 frames from the cell database. Each frame includes 2 cells,
and the input parameters were the same for all the methods.
The ground truth for the cells embryo database  only mentions  the center, thus a success was declared if we found at least one ellipse center within  20 from the ground truth center.
RANSAC, which treats noise as a possible part of the solution, achieved a 76\% success rate, against 100\% for our algorithm. 

\subsection{3D Experiments}

\begin{figure}
\begin{center}
\centerline{\includegraphics[width=5cm, height=2.5cm]{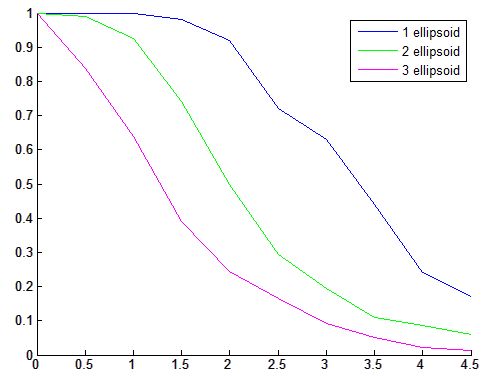}}
\end{center}
\caption{Fitting a noisy ellipsoid as a function of noise. The X-axis is the standard deviation of the gaussian noise. 
The Y-axis is the success fraction.}
\label{fig:3d_statistic}
\end{figure}

\begin{figure}[thb]
\begin{minipage}[b]{1.0\linewidth}
  \centering
  \centerline{\includegraphics[width=7cm,  height=4.5cm]{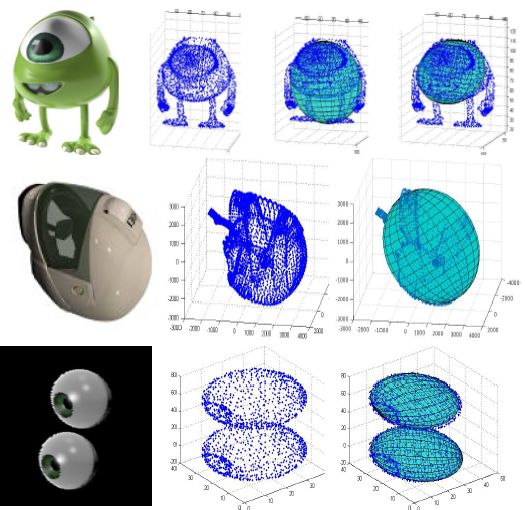}}
\end{minipage}
\caption{All the 3D examples have noise and/or multiple models}
\label{fig:models}
\end{figure}

To the best of our knowledge no published method fits multiple ellipsoids in three or more dimensions, so in this section we evaluate against synthetic ground truth, which is exact. We generated synthetic 3D models: three sets of 1000
3D models, containing one, two or three ellipsoids per model. The 3D models are in a box of $200^3$ with the semi-axes ranging from 20 to
60, orientation is uniformly distributed,
and ellipsoids can overlap. The points and tangents are perturbed by Gaussian noise. The mean-shift clustering threshold is set to 25 and the eigenvalue filtering threshold to 0.002. The outcomes for one, two and three ellipsoids are shown in Figure \ref{fig:3d_statistic}.
A success is counted when the parameters of the ellipsoid matrix --- the radii and angles together with the centre --- lie within 25 of the ground truth. 
Ellipsoid fitting for 3D models from the marketplace \cite{cgtrader} is shown in Figure \ref{fig:models}. 

The first image in each row is the rendered input  model, the second image is the  model points, the third image shows the output ellipsoids with the  cluster centers, and the fourth image shows the output ellipsoids surfaces. 

\subsection{Multidimensional}
To test the algorithm on multidimensional ellipsoids we generated 30 synthetic models for each dimension from 15 to 20, each containing one ellipsoid. The models lie in a range of $200^d$, with semi-axes from 20 to 60 and uniformly distributed orientation; the points and tangents are perturbed by Gaussian noise with standard deviation 1. As before we compared all output parameters, after normalisation, against the ground truth, requiring every parameter to lie within 10 of its true value. The success rate is 92.2\%.

\section{Conclusions}
We have presented an algorithm for detecting and fitting multiple ellipsoids that works in two, three and higher dimensions. Its two ingredients are a tangent-based linear formulation, in which each tangent contributes $d$ equations rather than one, and mean-shift clustering of normalised solutions in parameter space. The first reduces the number of samples each hypothesis requires and so reduces the number of iterations; the second makes the votes comparable, which is what allows multiple ellipsoids and heavy outlier contamination to be handled together.

The method outperforms the state of the art on real microscopy data in 2D, is more robust than RANSAC on the same inputs, and continues to recover ellipsoids correctly up to 20 dimensions, where we are not aware of a competing method.

Two limitations are worth stating. The method requires tangents as well as points, so it depends on an edge or surface estimate whose quality bounds the result; and the number of iterations, while far smaller than a points-only formulation would need, still grows with the number of ellipsoids and with the outlier fraction. Since the formulation is a robust solution of a homogeneous linear system, we expect it to apply to other quadric types and, more generally, to fitting problems beyond quadrics --- though we have not measured that here.

\section*{Acknowledgments}
This research was supported by the DFG, and  the Israel Science Foundation.

{\small
\bibliographystyle{abbrv}
\bibliography{refs}
}
\end{document}